\documentclass[11pt,a4paper]{article}
\usepackage[hyperref]{eacl2021}
\usepackage{times}
\usepackage{latexsym}
\usepackage{graphicx}
\usepackage[export]{adjustbox}

\usepackage{microtype}
\usepackage{enumitem}

\aclfinalcopy 

\usepackage{color, colortbl}
\definecolor{LightCyan}{rgb}{0.95,1,1}
\definecolor{LightYellow}{rgb}{1,1,0.88}
\definecolor{LightGrey}{rgb}{0.92,0.92,0.92}

\usepackage{subfig}

\usepackage{arabtex}
\usepackage{utf8}
\setcode{utf8}

\title{ArCorona: Analyzing Arabic Tweets in the Early Days of Coronavirus (COVID-19) Pandemic}

\author{Hamdy Mubarak, Sabit Hassan \\
  Qatar Computing Research Institute \\
  HBKU, Qatar \\
  \texttt{{\{hmubarak, sahassan2\}@hbku.edu.qa}}}

\date{}

\begin{document}
\maketitle
\begin{abstract}
Over the past few months, there were huge numbers of circulating tweets and discussions about Coronavirus (COVID-19) in the Arab region.  It is important for policy makers and many people to identify types of shared tweets to better understand public behavior, topics of interest, requests from governments, sources of tweets, etc. It is also crucial to prevent spreading of rumors and misinformation about the virus or bad cures. To this end, we present the largest manually annotated dataset of Arabic tweets related to COVID-19. We describe annotation guidelines, analyze our dataset and build effective machine learning and transformer based models for classification.
\end{list}
\end{abstract}

\section{Introduction}
\label{sec:introduction}

As the Coronavirus (COVID-19) crippled lives across the world, people turned to social media to share their thoughts, news about vaccines or cures, personal stories, etc. With Twitter being one of the popular social media platforms in the Arab region, tweets became a major medium of discussion about COVID-19. 
These tweets can be indicators of psychological and physical well being, public reactions to specific actions taken by the government and also public expectation from governments. Therefore, identifying types of tweets and understanding their content can aid decision making by governments. It is also important for governments to identify and prevent rumours and bad cures since they can bring harm to society. 

While there have been many recent works about tweets related to COVID-19, there are a very few targeted toward aiding governments in their decision making in the Arab region despite Arabic being one of the dominant languages on Twitter \cite{alshaabi2020a}. Some of the existing works use automatically collected datasets \cite{alqurashi2020large}. Manually labeled datasets are either small in size (few hundred tweets) \cite{alam2020fighting} or target a different task such as sentiment analysis \cite{haouari2020arcov19}. To fill this gap, we present and publicly share the largest (to our best knowledge) manually annotated dataset of Arabic tweets collected from early days of COVID-19, labeled for 13 classes.
We present our data collection and annotation scheme followed by data analysis, identifying trends, topics and distribution across countries. Lastly, we employ machine learning and transformer models for classification.
\section{Related Work}
\label{sec:related_work}
Much of recent works on COVID-19 rely on queries to Twitter or distant supervision. This allows a large number of tweets to be collected. \newcite{chen2020tracking} collect 123M tweets by following certain queries and accounts on Twitter. GeoCoV19 \cite{Qazi_2020} is a large-scale dataset containing 524M tweets with their location information. \newcite{banda2020largescale} collected 152M tweets at the time of their writing. \newcite{LI2020Characterizing} identifies situational information about COVID-19 and its propagation on Weibo. Other works include propagation of misinformation \cite{huang2020disinformation,shahi2020exploratory}, cultural, social and political impact of misinformation \cite{leng2020analysis} and rumor amplification \cite{cinelli2020covid19}. 

For Arabic, we see a similar trend where few datasets are manually labeled. \newcite{alqurashi2020large} provide a large dataset of Arabic tweets containing keywords related to COVID-19. Similarly, ArCOV-19 \cite{haouari2020arcov19} is a dataset of 750K tweets obtained by querying Twitter. \newcite{alam2020fighting} annotate a small number of English (currently 504) and Arabic tweets (currently 218) for (i) existence of claim and worthiness of fact-checking (ii) harmfulness to society, and (iii) relevance to governments or policy makers. \newcite{yang2020senwave} annotate 10K Arabic and English tweets for the task of fine-grained sentiment analysis. 

\newcite{alsudias-rayson-2020-covid} collected 1M unique Arabic tweets related to COVID-19 between December 2019 and April 2020. They used K-means algorithm from Scikit-learn Python package to cluster tweets into 5 clusters, namely: statistics, prayers, disease locations, advising, and advertising. They also annotated random 2000 tweets for rumor detection based on the tweets posted by the Ministry of Health in Saudi Arabia.

\section{Data Collection}

We used twarc search API\footnote{https://github.com/DocNow/twarc} to collect tweets having the Arabic word \<كورونا> (Corona) in Feb and March 2020. We collected 30M tweets in total. The reason behind selecting this word is that it's widely used by normal people, news media\footnote{https://www.aljazeera.com/topics/events/coronavirus-outbreak.html} and official organizations \footnote{https://www.who.int/ar/emergencies/diseases/novel-coronavirus-2019} as opposed to 19-\<كوفيد> (COVID-19) which is rarely used by normal people in different Arab countries based on our observations. We aimed to increase diversity of tweet sources. Our collection covers the period from Feb 21 until March 31 in which Coronavirus was reported for the first time in All Arab countries except United Arab Emirates (AE)\footnote{We use ISO 3166-1 alpha-2 for country codes}
(Jan 29) and Egypt (EG) (Feb 14).
The date of the first reported Coronavirus case in Lebanon (LB) was Feb 21, and in Iraq (IQ), Bahrain (BH), Oman (OM), and Kuwait (KW) was Feb 24, in Qatar was Feb 29, and in Saudi Arabia (SA) was March 2. All other Arab countries came later.

\section{Data Annotation}
\label{sec:data_annotation}

\begin{table*}[!h]
\small
\centering
\begin{tabular}{llr}
\hline \hline
\textbf{Class} & \textbf{Description} &  \textbf{Count}\\ \hline
\textbf{01. REP} & Reports and announcements such as number of infections, recovery cases and deaths. & 1664\\
\rowcolor{LightYellow}
\textbf{02. ACT} & Measures or actions taken by governments such as curfew, closing of country borders, shops & 1383\\
\rowcolor{LightYellow}
& and worship places. This includes discussions and consequences of these measures. & \\

\textbf{03. INFO} & Information about the virus, symptoms, incubation period, how it spreads, mask types, etc. & 300\\

\rowcolor{LightYellow}
\textbf{04. RUMOR} & Rumor or refute rumor. A rumor is a circulating story or report of uncertain or doubtful truth. & 421 \\

\textbf{05. ADVICE} & Advice or caution such as washing hands, staying at home, wearing masks and avoiding travel. & 1047 \\

\rowcolor{LightYellow}
\textbf{06. SEEK\_ACT} & Seek actions from governments such as closing airports, and  controlling prices of goods. & 587\\

\textbf{07. CURE} & News about good and bad cure, diagnosis, ventilators,  supportive medical equipment, etc. & 116\\

\rowcolor{LightYellow}
\textbf{08. VOLUNT} & Volunteering efforts or donation of money, goods or services. & 133\\

& & \\ \hline
\textbf{09. PRSNL} & Personal story or opinion. & 453\\
\rowcolor{LightYellow}
\textbf{10. SUPPORT} & Support or praise governments,  medical staff, celebrities, etc. & 386\\

\textbf{11. PRAYER} & Prayer. & 563\\
\rowcolor{LightYellow}
\textbf{12. UNIMP} & Unrelated or unimportant such as spams or advertisements. & 786\\

\textbf{13. NOT\_ARB} & Not Arabic, e.g. Persian. & 161\\
\hline
Total & & \textbf{8000}\\
\hline \hline
\end{tabular}
\caption{Annotation classes and distribution: Important classes (top) and LessImportant classes (bottom)}
\label{tab:classes}
\end{table*}

\begin{figure}[!h]
\begin{center}
\includegraphics[scale=0.60, frame]{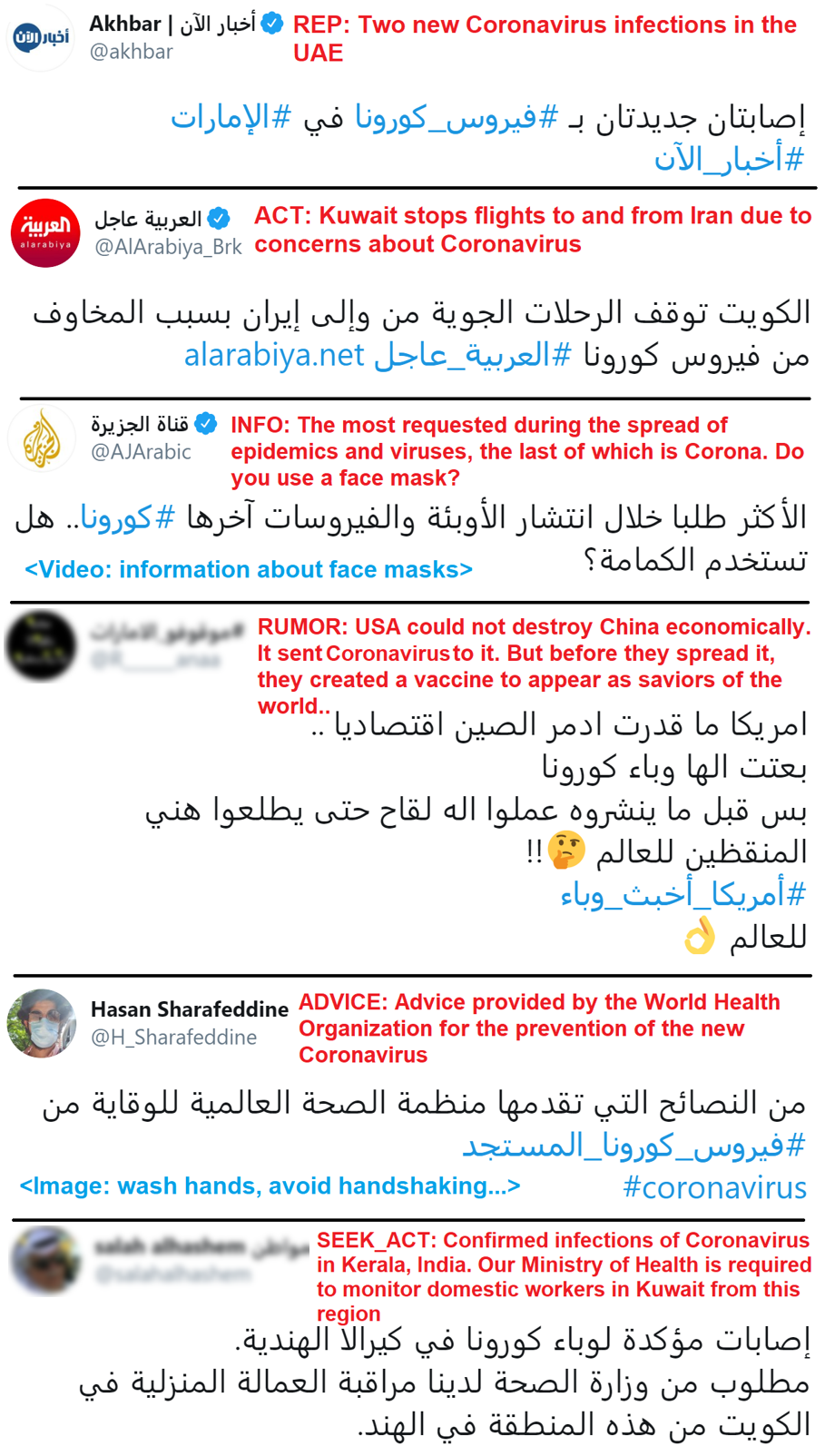} 
\caption{Examples for REP, ACT, INFO, RUMOR, ADVICE and SEEK\_ACT classes}
\label{fig:examples1}
\end{center}
\end{figure}

\begin{figure}[!h]
\begin{center}
\includegraphics[scale=0.60, frame]{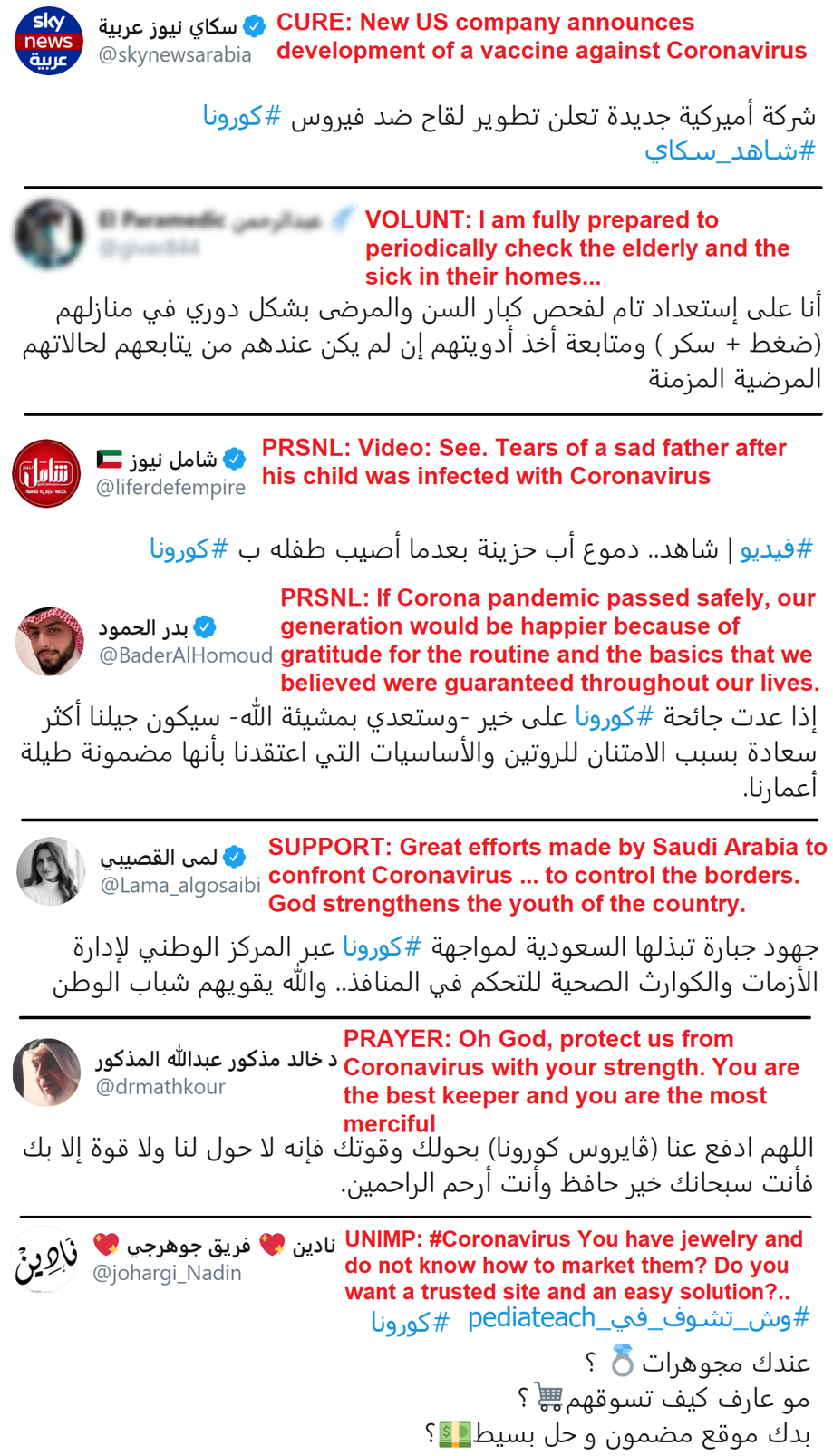} 
\caption{Examples for CURE, VOLUNT, PRSNL, SUPPORT, PRAYER and UNIMP classes}
\label{fig:examples2}
\end{center}
\end{figure}

During the period of our study (40 days), we extracted the top retweeted 200 tweets in each day (total of 8000). We assume that the top retweeted tweets are the most important ones which get highest attention from Twitter users. Annotation was done manually by a native speaker who is familiar with Arabic dialects according to class descriptions shown in Table \ref{tab:classes}. To measure quality, we annotated 200 random tweets by a second annotator. Inter-annotator agreement was 0.85 using Cohen’s kappa coefficient which indicates high quality given that annotation is not trivial and some classes are close to each other.  Examples of annotation classes are shown in Figures \ref{fig:examples1} and \ref{fig:examples2}.\\

\textbf{Note:} If a tweet has multimedia (image or video) or an external link (URL or another tweet), the annotator was asked to open it and judge accordingly to consider the full context of the tweet. For example, if a tweet has a text about a prayer and the attached image is about number of new cases, this should be classified as REP not PRAYER.\\
Data can be downloaded from this link\footnote{We share tweet id, date and class.}:\\ \url{https://alt.qcri.org/resources/ArCorona.tsv}.

\subsection{Limitation}
We found that $\approx$10\% of the tweets can take more than one class, e.g. a tweet reports new cases and a medical advice. We plan to allow multiple labels in future. In the current version, such tweets take the label of the first ``important" class. We consider the first 8 classes in Table \ref{tab:classes}  to be important and the last 5 classes (PRSNL, SUPPORT, PRAYER, UNIMP and NOT\_ARB) to be less important. These classes will be merged into LessImportant class.

\label{sec:data_collection}

\section{Analysis}
\label{sec:analysis}

\begin{figure*}[!h]
\begin{center}
\includegraphics[scale=0.80, frame]{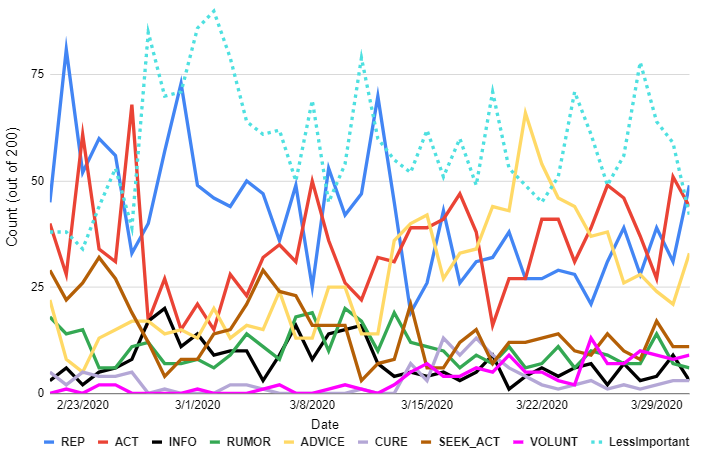} 
\caption{Timeline: Number of class tweets each day}
\label{fig:class-timeline}
\end{center}
\end{figure*}

\begin{figure}[!h]
\begin{center}
\includegraphics[scale=0.40, frame]{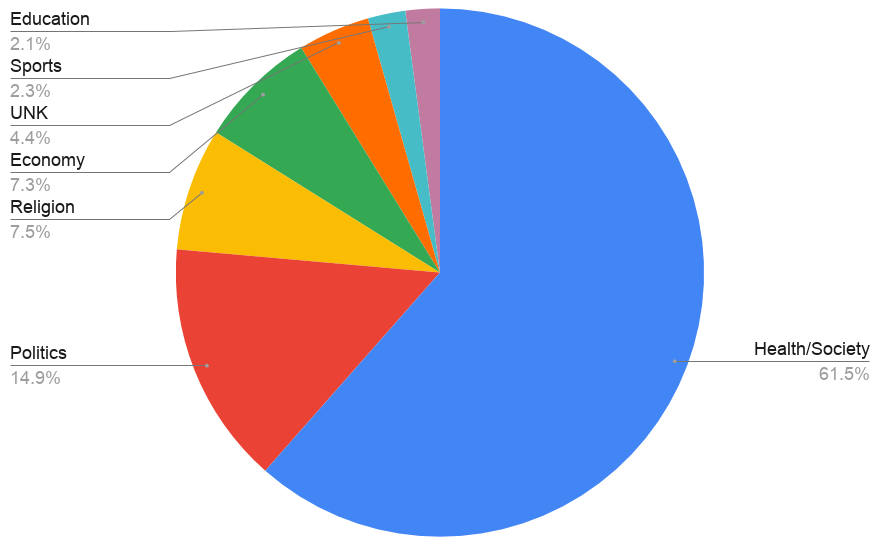} 
\caption{Tweet topics}
\label{fig:topics}
\end{center}
\end{figure}

\begin{figure}[!h]
\begin{center}
\includegraphics[scale=0.3, frame]{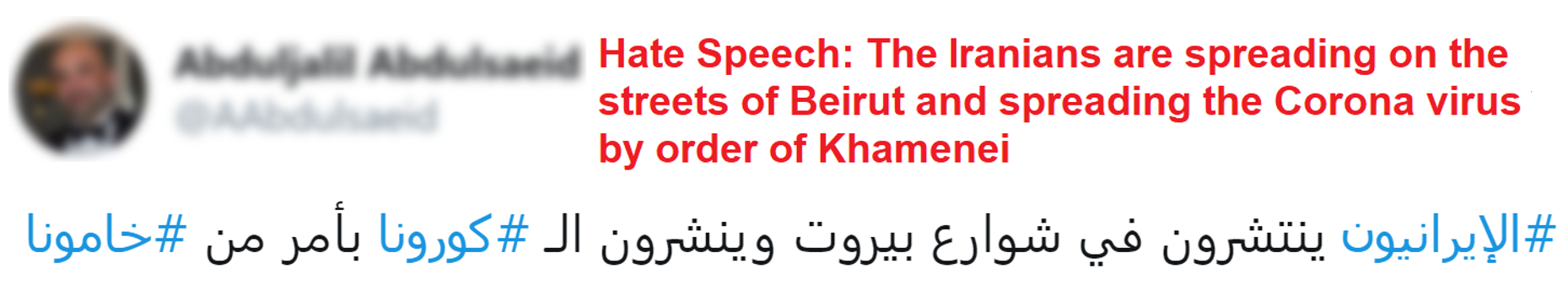} 
\caption{Hate speech example}
\label{fig:hate-speech}
\end{center}
\end{figure}

Class timeline is shown in Figure \ref{fig:class-timeline}. We can observe the following important notes:
\begin{itemize}[leftmargin=0.1in]
\setlength\itemsep{0em}
\item Large portion of tweets can be considered as LessImportant to many people ($\approx$ 30\%).
\item Reports (REP) and actions taken by governments (ACT) are the most retweeted tweets.
\item Information about the virus (INFO) get less attention with time and there is an increasing number of tweets about volunteering (VOLUNT).
\item There are continuous requests for governments to take actions (SEEK\_ACT) -- especially in the beginning ($\approx$ 15\%), and few tweets are about rumors ($\approx$ 5\%) and cures ($\approx$ 2\%).
\end{itemize}
We took a random sample of 1000 tweets and annotated them for their topics. Figure \ref{fig:topics} shows that, in addition to health, the virus affected many aspects of people's lives such as politics, economy, education, etc. We found also that 7\% of tweets have hate speech, e.g. attacking China and Iran for spreading the virus as shown in Figure \ref{fig:hate-speech}.

Table \ref{tab:country} shows country distribution and top accounts for the original authors of tweets.

Typically, people retweet tweets from ministry of health in their countries in addition to famous news agencies and celebrities. Most of these accounts are verified.

\begin{table}
\small
\centering
\begin{tabular}{lrl}
\hline
\textbf{Country} & \textbf{\%} &  \textbf{Top Accounts}\\ \hline
\rowcolor{LightYellow}
\textbf{SA} & 59  & SaudiNews50, SaudiMOH\\
\textbf{OTH} & 13  & MohamadAhwaze (SE), amjadt25 (UK)\\
\rowcolor{LightYellow}
\textbf{OM} & 7  & OmanVSCovid19, OmaniMOH\\
\textbf{KW} & 7  & Almajlliss, KUWAIT\_MOH\\
\rowcolor{LightYellow}
\textbf{QA} & 4  & amansouraja, MOPHQatar\\
\textbf{AE} & 4  & AlHadath, AlArabiya\_Brk\\
\rowcolor{LightYellow}
\textbf{EG} & 3  & RassdNewsN, mohpegypt\\
\end{tabular}
\caption{Country distribution and top accounts}
\label{tab:country}
\end{table}

\section{Experiments and Evaluation}
We randomly split the data into sets of 6000, 1000 and 1000 tweets for train, dev and test sets respectively. We report macro-averaged Precision (P), Recall (R) and F1 score along with Accuracy (Acc) on test set \footnote{Differences between dev and test sets are $\pm 2-3\%$ (F1).}. We use F1 score as primary metric for comparison.
\label{sec:experiments_results}
\subsection{Features}
\paragraph{N-gram features} We experimented with character and word n-gram features weighted by term frequency-inverse term document frequency (tf-idf). We report results for only the most significant ranges, namely, word [1-2] and character [2-5].
\paragraph{Mazajak Embeddings} Mazajak embeddings are word-level skip-gram embeddings trained on 250M Arabic tweets, yielding 300-dimensional vectors \cite{abu-farha-magdy-2019-mazajak}.

\subsection{Classification Models}
\paragraph{Support Vector Machines (SVMs)} SVMs have been shown to perform decently for Arabic text classification tasks such as spam detection \cite{spamMubarak2020}, adult content detection \cite{mubarak2021adult,hassan2021asad}, offensiveness detection \cite{hassan-etal-2020-alt,hassan-etal-2020-alt-semeval} or dialect identification \cite{abdelali2020arabic,bouamor-etal-2019-madar}.
We experimented with i) word n-gram, ii) character n-gram and iii) Mazajak Embeddings. We used LinearSVC implementation by scikit-learn \footnote{https://scikit-learn.org/}.
\paragraph{Deep Contextualized Transformer Models (BERT)}
Transformer-based pre-trained contextual embeddings, such as BERT \cite{devlin-etal-2019-bert}, have outperformed other classifiers in many NLP tasks. We used AraBERT \cite{Antoun2020AraBERTTM}, a BERT-based model trained on Arabic news. 
We used ktrain library \cite{maiya2020ktrain} that utilizes Huggingface\footnote{https://huggingface.co/} implementation to fine-tune AraBERT. We used learning rate of 8e\textsuperscript{-5}, truncating length of 50 and fine-tuned for 5 epochs. 


\subsection{Binary Classification}
First, we experiment to distinguish LessImportant tweets from others (see Section \ref{sec:data_annotation}). From Table \ref{tab:2-way}, we can see that SVMs with character [2-5]-gram outperformed others with F1 score of \textbf{79.8}, closely followed by AraBERT with \textbf{79.6} F1.
\begin{table}
\small
\centering
\begin{tabular}{cccccc}
\hline \textbf{Model} & \textbf{Features} & \textbf{Acc.} &\textbf{P} & \textbf{R} & \textbf{F1}\\ \hline
Majority Class & -    &   72.5    &	36.3    &	50.0    &	42.0\\
\rowcolor{LightGrey}
SVM & W[1-2]    & 84.4	&   82.5	&   76.4   &	78.6\\
SVM & C[2-5]    & \textbf{85.4}	&   \textbf{84.3}   &	77.4   &	\textbf{79.8}\\
\rowcolor{LightGrey}
SVM & Mazajak   &  83.9   &	80.5    &	77.7    &	78.9\\
AraBERT &       &       83.9  &	80.0    &	\textbf{79.2}    &	79.6\\
\hline
\end{tabular}
\caption{\label{tab:2-way} Binary classification results}
\end{table}

\subsection{Fine-grained Classification}
\begin{table}
\small
\centering
\begin{tabular}{cccccc}
\hline \textbf{Model} & \textbf{Features} & \textbf{Acc.} &\textbf{P} & \textbf{R} & \textbf{F1}\\ \hline
Majority Class & -    &   22.7    &	1.7 &	7.7 &	2.8\\
\rowcolor{LightGrey}
SVM & W[1-2]    &   \textbf{62.8}   &	\textbf{64.3}   &	53.5   &	56.3\\
SVM & C[2-5]    &   59.0	&   \textbf{64.3}   &	49.4   &	51.8\\
\rowcolor{LightGrey}
SVM & Mazajak   &  60.0    &	55.1    &	51.5    &	52.4\\
AraBERT &       &      62.7   &	61.6    &	\textbf{59.8}    &	\textbf{60.5}\\
\hline
\end{tabular}
\caption{\label{tab:13-way} Fine-grained classification results}
\end{table}
Our next set of experiments were designed for fine-grained classification for 13 classes. With F1 score of \textbf{60.5}, AraBERT outperformed others (Table \ref{tab:13-way}).\\

\begin{figure}[!h]
\begin{center}
\includegraphics[scale=0.18, frame]{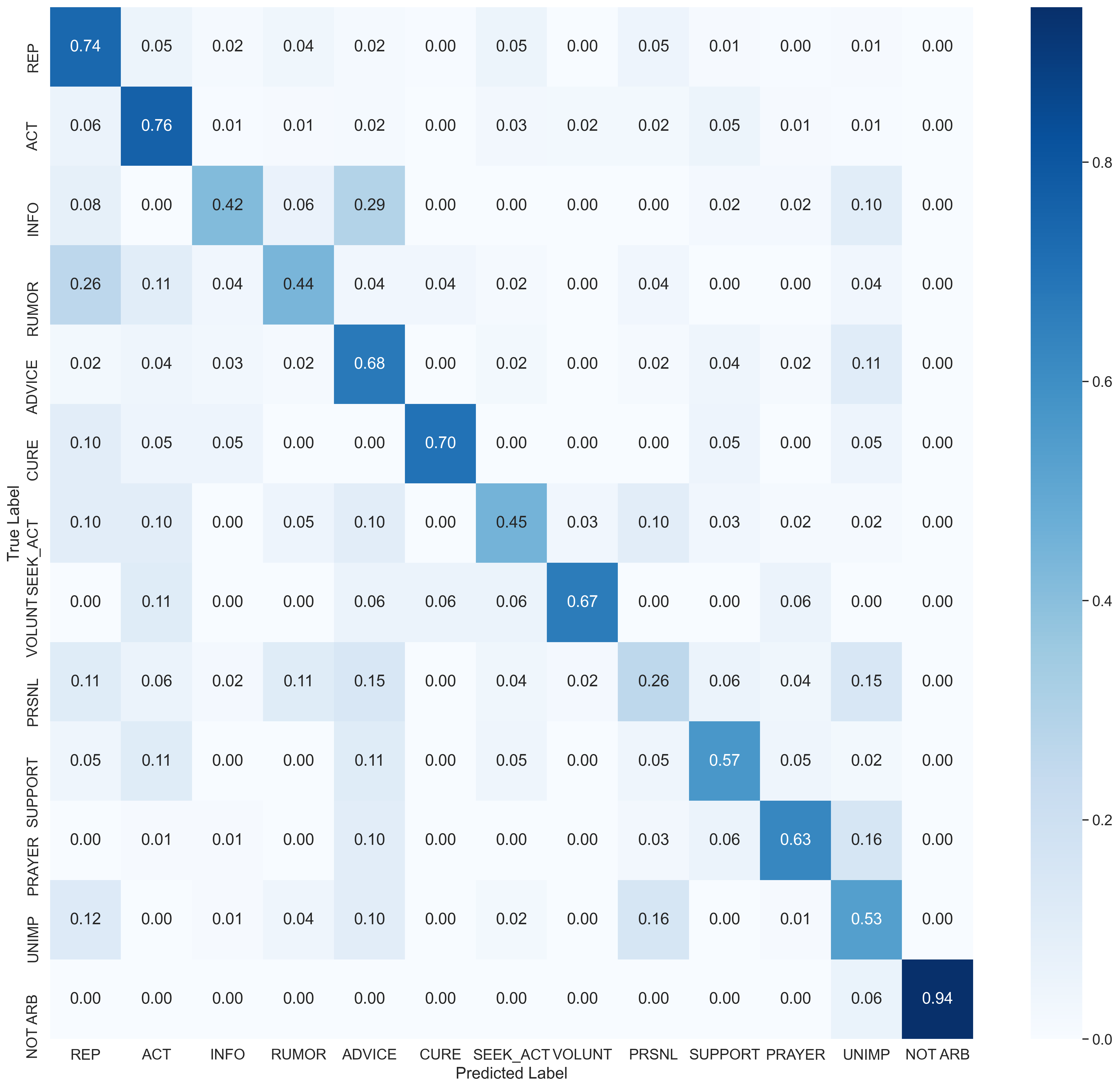} 
\caption{Confusion matrix normalized over true labels}
\label{fig:conf}
\end{center}
\end{figure}

\textbf{Error Analysis: }
AraBERT confusion matrix (Figure  \ref{fig:conf}) shows that PRSNL, INFO and RUMOR are the hardest classes to identify and the most common error is misclassifying INFO as ADVICE.
We hypothesize these errors can be reduced if larger data set is being annotated.

\section{Conclusion and Future Work}
\label{sec:conclutions}
We present the largest publicly available manually annotated dataset of Arabic tweets for 13 classes that includes the most retweeted  tweets in the early days of COVID-19. Followed by data analysis, we present models that can reliably identify important tweets and can perform fine-grained classification. In the future, we plan to compare our data to data from later days of the pandemic.

\bibliography{anthology,eacl2021}
\bibliographystyle{acl_natbib}

\end{document}